%% file: main.tex
\documentclass[11pt,onecolumn]{cleantechnicalreport}
\usepackage{array,makecell,placeins,eso-pic}
\usepackage[authoryear,round]{natbib}
\usepackage{tactalent-layout}
\hypersetup{pdftitle={GPT-6 Astra Is a Native Visuo-Tactile Policy for Contact-Rich Manipulation},pdfauthor={Chen Yang, Linzhe Shi, Hang Zhang, Mu Xu, Jiansheng Fan, Chen Wang}}
\newcommand{\V}{\textsc{V}}
\newcommand{\VT}{\textsc{V+T}}
\newcommand{\record}[1]{\href{https://tactalent.github.io/gallery.html?lang=en\&episode=#1}{Full rollout}}
\begin{document}
\normalsize\raggedbottom
\thispagestyle{plain}
% Institution marks share an optical centre line and align to the text edges.
\AddToShipoutPictureFG*{%
  \AtPageUpperLeft{%
    \put(\LenToUnit{20mm},\LenToUnit{-18mm}){%
      \includegraphics[height=12mm]{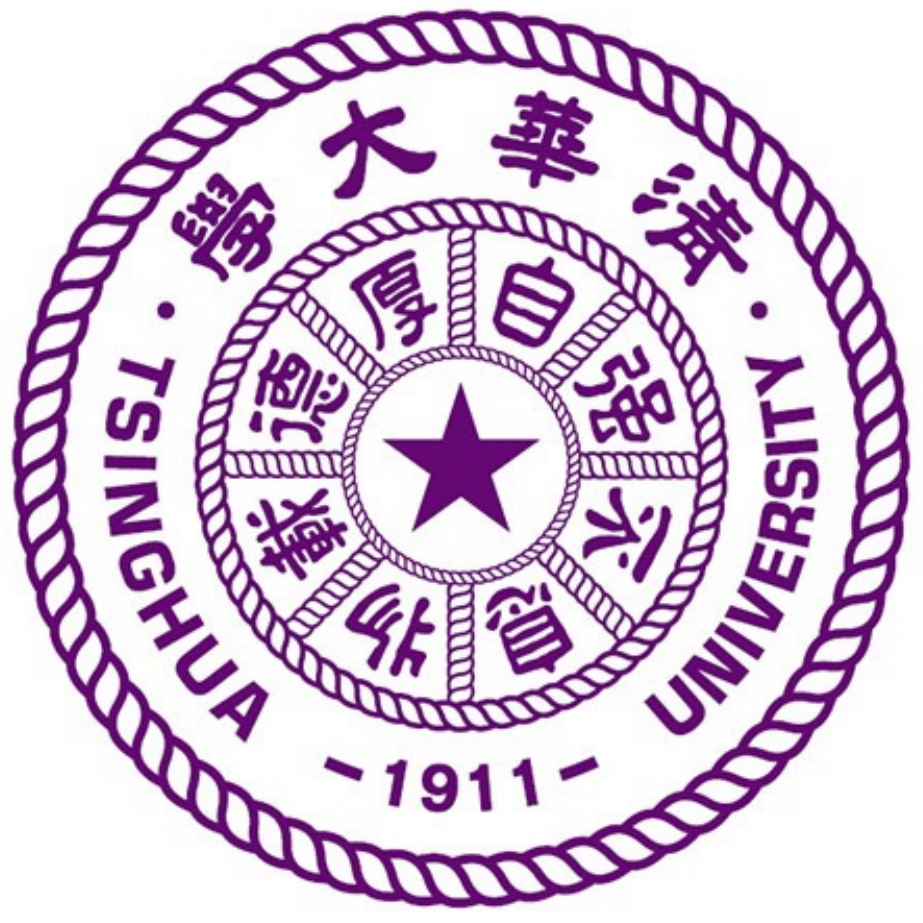}}%
    \put(\LenToUnit{190mm},\LenToUnit{-16.5mm}){%
      \makebox[0pt][r]{\includegraphics[height=9mm]{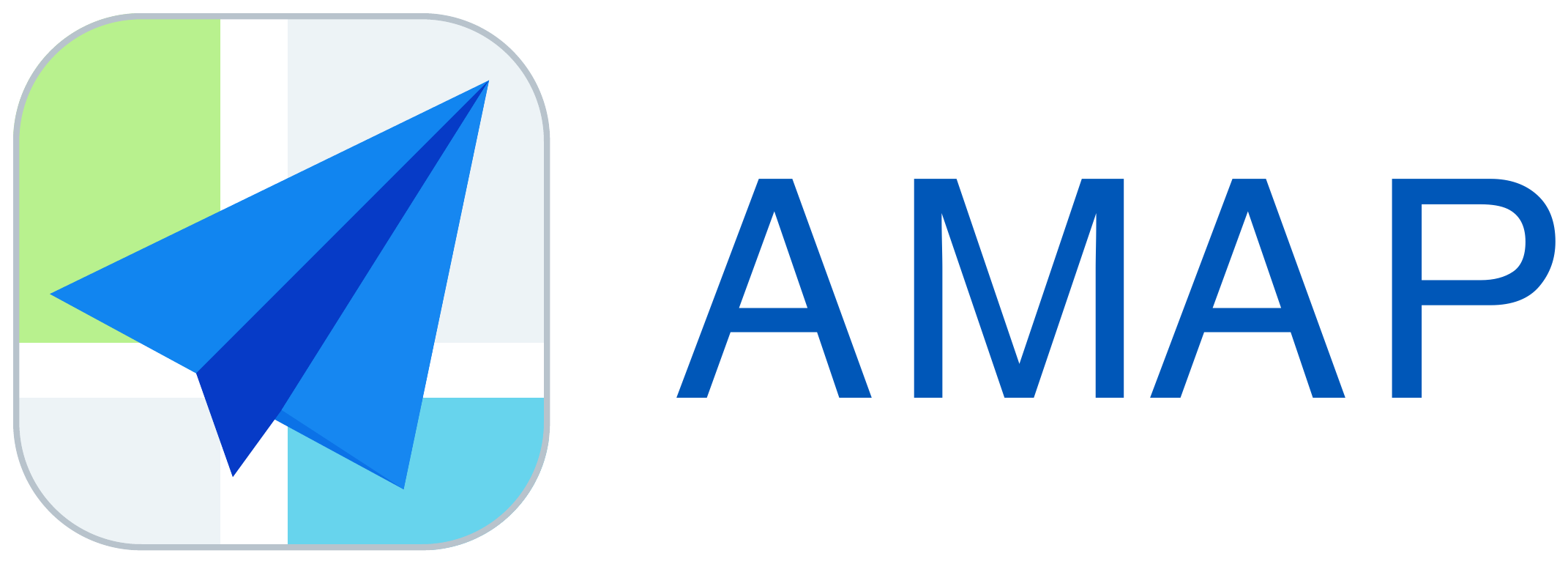}}}}}
\noindent\textcolor{Accent}{\rule{\linewidth}{1.7pt}}\par
\vspace{12pt}
{\raggedright\sffamily\bfseries\fontsize{20}{24}\selectfont\reporttitlefont
GPT-6 Astra Is a Native Visuo-Tactile Policy\\[1pt]
for Contact-Rich Manipulation\par}
\vspace{12pt}
{\raggedright\sffamily\bfseries\fontsize{10.2}{14}\selectfont
Chen Yang\textsuperscript{1,2}\quad Linzhe Shi\textsuperscript{2}\quad Hang Zhang\textsuperscript{2}\quad
Mu Xu\textsuperscript{2}\quad Jiansheng Fan\textsuperscript{1}\quad Chen Wang\textsuperscript{1,\dag}\par}
\vspace{5pt}
{\raggedright\fontsize{10}{13}\selectfont\color{Muted}
\textsuperscript{1}Tsinghua University\qquad\textsuperscript{2}Amap, Alibaba Group\par
\textsuperscript{\dag}Corresponding author\par}
\vspace{11pt}
\begin{reportabstract}
Recent advances in multimodal models have enabled GPT-6 Astra to perform robotic manipulation through in-context learning. Yet contact-rich control and tactile perception are often regarded as strengths of specialist policies, leaving open whether Astra can directly understand and use tactile observations. We provide, to our knowledge, the first demonstration that Astra directly interprets vision-based tactile images and uses their contact cues for manipulation. Tactile RGB images enter through its existing visual interface, enabling few-shot closed-loop control on nineteen simulated UniVTAC and OpenViTac tasks with 210 sampled scene seeds per setting. With visuo-tactile input, Astra achieves mean success rates of 93.8\% and 81.8\%, exceeding the strongest reported specialist baselines in our comparison by 22.9 and 13.1 percentage points, respectively. Vision-only performance also leads these baselines, reaching 76.3\% and 80.6\%. Trajectory analysis reveals active probing, cautious grasping, contact-guided alignment, and recovery, alongside difficulties with tangential loading and corrective execution. Our findings challenge the exclusivity of specialist policies in contact-rich manipulation and show that fine-grained vision can partially compensate for missing touch. They also identify vision-based tactile sensing as an interface for scaling contact understanding with foundation models and reveal signs of benchmark saturation. These results motivate joint visual--tactile pretraining, harder contact-rich evaluations, and physical-robot validation.
\end{reportabstract}
\vspace{2pt}
{\fontsize{9}{12}\selectfont\urlstyle{tt}
\noindent{\sffamily\bfseries Date:}\enspace October 20, 2026\\[2pt]
{\sffamily\bfseries Code:}\enspace\url{https://github.com/ccyydd/TacTalent}\\[2pt]
{\sffamily\bfseries Project Page:}\enspace\url{https://tactalent.github.io/}\\[2pt]
{\sffamily\bfseries Video Gallery:}\enspace\url{https://tactalent.github.io/gallery.html}\par}
\vspace{10pt}
\begin{figure}[!htbp]\centering\includegraphics[width=\linewidth]{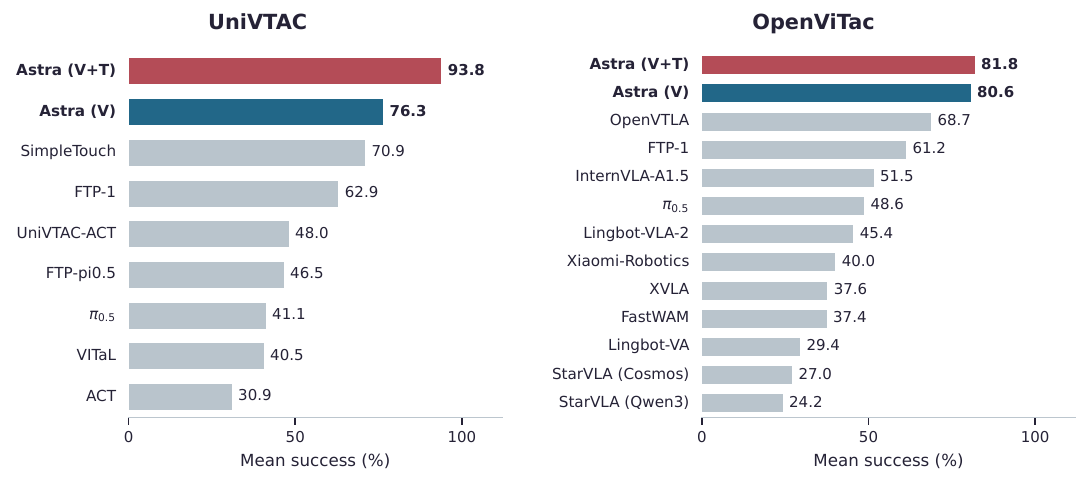}\caption{\textbf{Benchmark-level results.} Task-averaged success rates (\%). Astra uses vision alone (blue) or vision with tactile RGB images (red). Grey bars show published specialist baselines. Models are ordered by their reported mean within each benchmark. Task-level results and evaluation scope appear in Section~\ref{sec:results}.}\label{fig:overview}\end{figure}

\clearpage

\section{Introduction}\label{sec:intro}
Recent studies show that GPT-6 Astra can translate visual observations and instructions into robot actions, challenging the assumption that effective manipulation requires a separately trained policy. GPT-as-Policy~\citep{su2026astra}, RoboProbe~\citep{zhang2026roboprobe}, RoboDawn~\citep{guo2026robodawn}, and RoboICL~\citep{liu2026roboicl} explore this capability through different control interfaces and forms of contextual feedback. GPT-Policy and broader embodied evaluations further examine how demonstration following and task decisions translate into physical execution~\citep{cheng2026gptpolicy,galbot2026systematic}. These advances shift the question from whether general models can control robots to which capabilities still distinguish specialist policies.

Contact-rich manipulation is a prominent candidate. Precise insertion, stable grasping, and material discrimination depend on local geometry, slip, and physical properties that scene images may not reveal. Motions that work in free space can fail after contact, while a stronger grasp can damage a fragile object. These demands motivate tactile pretraining and learned visual--tactile integration, including FTP-1~\citep{yuan2026ftp}, OpenVTLA~\citep{wu2026openvitac}, and SimpleTouch~\citep{yang2026simpletouch}. Can Astra directly use contact evidence for these decisions, or do such interactions require specialist training?

Vision-based tactile sensing makes this question especially relevant. Sensors such as GelSight encode contact geometry and deformation as images~\citep{yuan2017gelsight}, exposing touch through the same interface that supports general visual understanding. Few-shot demonstrations relate local deformation to preceding actions, grasp conditions, and resulting object motion. We explore whether Astra can interpret these relationships and use tactile evidence to guide a new interaction, extending visual competence to contact understanding.

We investigate Astra's direct tactile-image understanding on UniVTAC~\citep{chen2026univtac} and OpenViTac~\citep{wu2026openvitac}. Tactile RGB images enter through the model's existing visual interface, allowing us to compare vision-only and visuo-tactile control on the same sampled scenes. We connect task outcomes to recorded contact inference, corrective actions, and recovery. Our contributions are threefold:
\begin{enumerate}[leftmargin=16pt,itemsep=4pt,topsep=5pt]
\item \textbf{Direct tactile-image understanding and control.} To our knowledge, we provide the first demonstration of Astra directly interpreting vision-based tactile images and using their contact cues for manipulation. We investigate this capability on nineteen tasks and 210 sampled scene seeds per input setting.
\item \textbf{Leading contact-rich manipulation results.} Astra reaches 93.8\% and 81.8\% mean success, exceeding the strongest reported specialist means in our comparison by 22.9 and 13.1 percentage points (Figure~\ref{fig:overview}). It outperforms FTP-1 and SimpleTouch on UniVTAC and OpenVTLA on OpenViTac; vision-only performance also leads these baselines at 76.3\% and 80.6\%.
\item \textbf{An analysis of contact understanding and visual compensation.} We connect task-level outcomes to contact-guided probing, fine adjustment, and recovery, and identify subtle visual cues that compensate for missing touch. This analysis reveals both Astra's tactile competence and its remaining perceptual and control limitations.
\end{enumerate}

Together, these results challenge the presumed moat of specialist contact-rich policies and connect tactile sensing to the development of more capable general models.

\section{Related Work}\label{sec:related}
\paragraph{Multimodal Models as Robot Policies.}
Code as Policies and VoxPoser use general models to generate robot programs and spatial objectives~\citep{liang2023codepolicies,huang2023voxposer}. Recent Astra-based work brings the model closer to direct control: GPT-as-Policy compares direct and hybrid policies~\citep{su2026astra}; RoboProbe and systematic evaluations examine manipulation and execution limits~\citep{zhang2026roboprobe,galbot2026systematic}; and RoboDawn and GPT-Policy ground control in demonstrations~\citep{guo2026robodawn,cheng2026gptpolicy}. RoboICL unifies demonstrations and online experience through an observation--action--receipt--observation grammar, preserving distributed interactions alongside recent feedback~\citep{liu2026roboicl}. Tactile feedback has also been studied in Astra-based peg insertion~\citep{tanaka2026taczero}. This line complements learned action-chunking, diffusion, and pretrained VLA policies~\citep{zhao2023act,chi2023diffusionpolicy,kim2025openvla,intelligence2025pi05}. Our study centers on Astra's direct interpretation of tactile RGB images and their role in contact-rich control.

\paragraph{End-to-End Policies with Tactile Feedback.}
VITaL, 3D-ViTac, and ViTacFormer learn visual--tactile representations for manipulation~\citep{george2024vital,huang2024threedvitac,heng2025vitacformer}. Tactile-VLA, VLA-Touch, and AT-VLA incorporate tactile feedback into pretrained VLAs~\citep{huang2025tactilevla,bi2025vlatouch,li2026atvla}. FTP-1 pretrains tactile policies across sensors~\citep{yuan2026ftp}; OpenVTLA studies tactile representations and integration within OpenViTac~\citep{wu2026openvitac}; and SimpleTouch adds a tactile expert with predictive contact features, without additional tactile policy pretraining~\citep{yang2026simpletouch}. TacDyn-WAM models implicit tactile dynamics by predicting future representations within a heterogeneous visuo-tactile world action model~\citep{wang2026tacdynwam}. Sparsh, AnyTouch, and T3 provide reusable sensory representations across sensing configurations and tasks~\citep{higuera2024sparsh,feng2025anytouch,zhao2025t3}. We examine how much tactile competence a general model can access directly through its visual interface, using demonstrations that relate contact images to actions and outcomes.

\paragraph{Contact-Rich Manipulation Benchmarks.}
LIBERO, ManiSkill3, and RoboDojo evaluate manipulation learning, transfer, and generalist control~\citep{liu2023libero,tao2024maniskill3,chen2026robodojo}; Bench2Dex extends evaluation to visuo-tactile bimanual manipulation across dexterous hands~\citep{yang2026bench2dex}. UniVTAC unifies tactile data generation, learning, and benchmarking~\citep{chen2026univtac}, supporting established comparisons across grasping, classification, placement, and insertion~\citep{yuan2026ftp,yang2026simpletouch}. The more recent OpenViTac provides a complementary sim-and-real framework for property discrimination and precise contact manipulation~\citep{wu2026openvitac}. Our evaluation spans both suites and examines whether texture, contour, and visible slip can support nominally tactile judgments, helping distinguish fine-grained visual competence from understanding of genuinely unobservable physical properties.

\section{Method: Direct Tactile-Image Control}\label{sec:method}
\begin{figure}[!htbp]\centering\includegraphics[width=\linewidth]{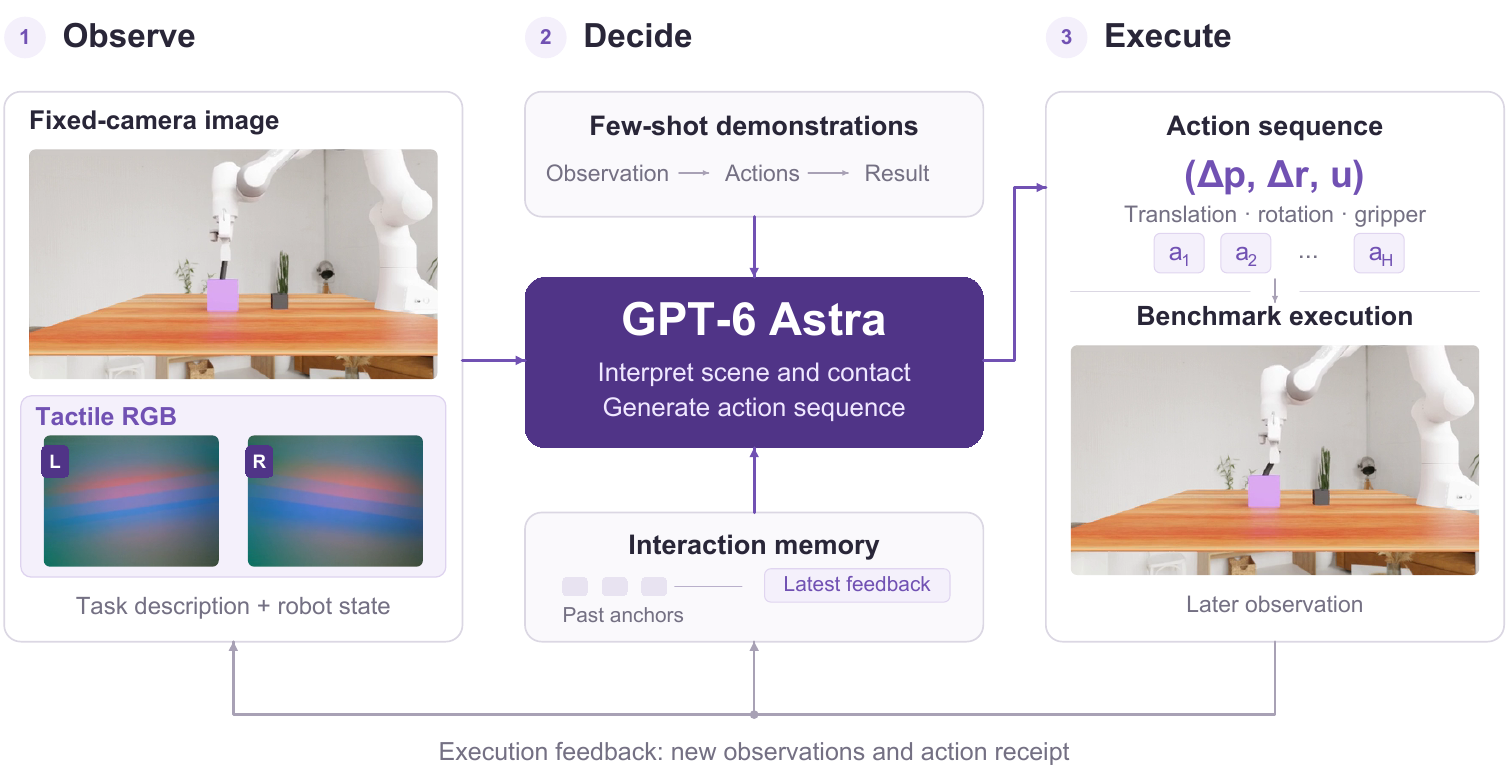}\caption{\textbf{Direct tactile-image input for contact-guided manipulation.} Tactile RGB images and scene observations enter Astra through its visual input interface. Demonstrations and online interaction memory connect contact evidence to actions and outcomes. Astra generates end-effector and gripper commands; benchmark execution returns observations and a receipt for the next decision. The displayed fixed-camera and tactile images are actual inputs from Insert Hole, seed 1000028. Vision-only evaluation omits the tactile images throughout the context.}\label{fig:method}\end{figure}

We present tactile images to Astra as direct visual evidence for contact-rich manipulation. Together with scene images and robot state, they expose local deformation and contact geometry through the model's existing visual interface. Demonstrations connect these observations to actions and outcomes, and online feedback supports contact-guided corrections. Figure~\ref{fig:method} shows this interface with an actual Insert Hole observation. Demonstration-conditioned execution and interaction memory follow RoboICL~\citep{liu2026roboicl}.

\subsection{Scene and Contact Observations}
At decision round $t$, the model receives task instruction $g$, robot state $x_t$, scene images $I_t$, and, in the visuo-tactile setting, the two finger-contact images $T_t$. Scene images describe global object pose and target geometry; tactile images describe local deformation and contact. The input preserves the benchmark's configured cameras, including a wrist view where available. Robot state provides the current end-effector pose and proprioceptive information needed to relate an observed contact to a proposed motion. Demonstration and online observations use the same image interface, allowing Astra to compare contacts in relation to their associated actions and outcomes. Vision-only control omits tactile images from both sources.

\subsection{Demonstrations and Interaction Memory}
The demonstration context $\mathcal{D}$ consists of selected blocks from successful task trajectories. A block contains its initial observation and robot state, the recorded action sequence, an execution receipt, and the resulting observation and state. Blocks are selected before evaluation and remain available throughout an episode. They expose local state transitions across the trajectory; tasks with distinct classes or shapes include demonstrations that cover those branches. Our preparation uses temporal targets constrained by action validity and non-overlap, as detailed in Appendix~\ref{app:configuration}.

Online memory $\mathcal{M}_t$ uses the same interaction structure. Distributed anchors retain earlier task stages, while the latest interaction exposes the most recent execution result. Explicit gap markers identify omitted intervals. This bounded memory relates the current observation to earlier attempts and demonstrated procedures~\citep{liu2026roboicl}.

\subsection{Action Generation and Feedback}
Astra produces an ordered sequence of end-effector and gripper commands from the assembled context:
\begin{equation}
A_t \sim \pi\!\left(\,\cdot\mid g,\mathcal{D},\mathcal{M}_t,x_t,I_t,T_t\right),
\qquad a=(\Delta p,\Delta r,u),
\label{eq:policy}
\end{equation}
where $\Delta p$ and $\Delta r$ specify translation and rotation increments and $u$ is the gripper target. In the vision-only setting, $T_t$ is absent. The benchmark adapter converts these commands to the robot's control interface and records their execution. The receipt and next observation complete the interaction used for the following decision.

Execution feedback distinguishes a completed command from achieved task progress: an insertion command can still leave an object blocked at the rim. New observations allow Astra to revise the next action. Short action notes accompany decisions and link commands to the model's assessment; our analysis combines these notes with the visible changes and task outcomes.

\section{Experimental Setup}\label{sec:setup}
\paragraph{Benchmarks and Sampling.}
We evaluate eight UniVTAC tasks and eleven OpenViTac tasks~\citep{chen2026univtac,wu2026openvitac}. UniVTAC supplies an established tactile-policy comparison across grasping, placement, classification, and precise insertion. The more recent OpenViTac adds property discrimination and contact-sensitive assembly within a paired sim-and-real framework. Across seven policies on eight shared tasks, its authors report strong agreement between simulated and real-world performance (Pearson $r=0.967$), supporting its use for comparative evaluation~\citep{wu2026openvitac}. We use a 10\% stratified sample, with sampling seed 42, preserving class and layout proportions. Each input setting contains 80 UniVTAC and 130 OpenViTac scene seeds: ten per task, except Insert Block v2, which contains ten per shape across three shapes.

\paragraph{Model and Demonstrations.}
The model is GPT-6 Astra with \texttt{xhigh} reasoning effort. We use UniVTAC's released training recordings~\citep{chen2026univtac} and collect successful OpenViTac demonstrations with its official scripted experts in training mode~\citep{wu2026openvitac}. Demonstrations provide context, while separate evaluation episodes determine success rates. Twelve tasks use one demonstration with twelve selected segments. Six classification or selection tasks use two demonstrations with five segments each; Insert Block v2 uses three demonstrations, one per shape, also with five segments each. Segment length $H$ is chosen with reference to trajectory length and fixed per task. Each segment pairs its $H$ actions with the initial and resulting observations. Appendix~\ref{app:configuration} gives every task's configuration and the selection procedure. All selected demonstration blocks remain available at every decision throughout the episode.

\paragraph{Input Conditions and Camera Access.}
Vision-only (\V) and visuo-tactile (\VT) settings use the same demonstration trajectories, segment starts, actions, and sampled evaluation scenes. \V\ excludes tactile images from both demonstrations and online observations; \VT\ additionally includes the left and right tactile RGB images. Camera selection strictly follows each benchmark's official task configuration. UniVTAC uses only the fixed camera in six tasks and also provides the wrist camera for Lift Can and Insert Tube. OpenViTac provides both fixed and wrist cameras for every task. All settings also receive task text and robot proprioception, and generate the seven-dimensional commands in Equation~\ref{eq:policy}.

\paragraph{Success Criteria.}
All tasks retain their official success criteria except Grasp Chip. Its original conditions impose unnecessarily strict execution details beyond intact object transfer. We simplify these conditions: an intact chip must be grasped, transferred, and placed with stable physical support in the target region. The adjusted criterion accepts supported placement without requiring a particular release-and-withdrawal sequence. The task still requires maintaining the chip's integrity during manipulation.

\paragraph{Baselines and Metrics.}
The tables compare Astra with published benchmark results, including action-chunking, VLA, world-action, and tactile-policy baselines~\citep{zhao2023act,george2024vital,intelligence2025pi05,yuan2026ftp,yang2026simpletouch,wu2026openvitac}. Astra is evaluated on the sampled scenes; baseline values are the corresponding reported benchmark results. We report per-task success percentages and equal-weight means across the eight or eleven tasks. The thirty Insert Block v2 scenes are aggregated into a single task score with the same weight as every other task. The task breakdown separately examines property judgments, constrained insertion, stable grasping, and transfer.

\section{Results: Five Findings}\label{sec:results}

\subsection{Leading Performance on Contact-Rich Tasks}\label{sec:result-performance}
With visual and tactile inputs, Astra reaches 93.8\% mean success on UniVTAC and 81.8\% on OpenViTac (Tables~\ref{tab:univtac} and~\ref{tab:openvitac}). The strongest reported baseline means in these tables are 70.9\% for SimpleTouch~\citep{yang2026simpletouch} and 68.7\% for OpenVTLA~\citep{wu2026openvitac}, yielding gaps of 22.9 and 13.1 percentage points. Vision-only Astra also exceeds these means, reaching 76.3\% and 80.6\%, respectively. These results establish general multimodal models as strong contact-rich manipulation policies, surpassing the reported specialist baselines in both benchmark comparisons.

Astra also exceeds FTP-1~\citep{yuan2026ftp} by 30.9 percentage points on UniVTAC and 20.6 points on OpenViTac. The improvements extend to precise insertion and fragile transfer, complementing the semantic and open-ended strengths identified by earlier Astra evaluations~\citep{zhang2026roboprobe,liu2026roboicl}. Weight Classify remains difficult, while several classification, transfer, and insertion tasks reach 100\% in the sampled episodes.

The task breakdown reveals that the gains span more than one operation family. On UniVTAC, the largest changes occur in constrained extraction and hole insertion, while the already strong Lift Bottle and Grasp Classify results remain unchanged. On OpenViTac, improvements in Hardness Classify and Insert Block v1 coexist with high vision-only scores on texture discrimination and several transfer tasks. This pattern motivates examining what evidence drives a particular decision, rather than interpreting touch as a uniform addition to every task.

\begin{table}[!htbp]\centering
\caption{\textbf{UniVTAC success rates (\%).} Baseline values are reported in UniVTAC and SimpleTouch~\citep{chen2026univtac,yang2026simpletouch}. Astra uses ten sampled scene seeds per task. \V: vision; \VT: vision and tactile RGB. Avg. is the unweighted mean over tasks. Bold denotes the best value in each column, including ties.}\label{tab:univtac}
\vspace{4pt}
{\fontsize{8.1}{10.6}\selectfont\setlength{\tabcolsep}{2pt}\renewcommand{\arraystretch}{1.13}\input{table-univtac.tex}}
\end{table}
\vspace{5pt}
\begin{table}[!htbp]\centering
\caption{\textbf{OpenViTac success rates (\%).} Baseline values are reported in OpenViTac~\citep{wu2026openvitac}. Astra uses ten sampled scene seeds per task, except Insert Block v2 (thirty across three shapes). \V: vision; \VT: vision and tactile RGB. Avg. is the unweighted mean over tasks. Bold denotes the best value in each column, including ties. Class. abbreviates Classify; Roughness Regrasp abbreviates Roughness-Guided Regrasp. Grasp Chip follows the adjusted criterion described above.}\label{tab:openvitac}
\vspace{4pt}
{\fontsize{7.25}{9.6}\selectfont\setlength{\tabcolsep}{1.0pt}\renewcommand{\arraystretch}{1.16}\input{table-openvitac.tex}}
\end{table}

\subsection{Direct Tactile Understanding and Control}\label{sec:tactile}
\begin{figure}[!htbp]\centering\includegraphics[width=\linewidth]{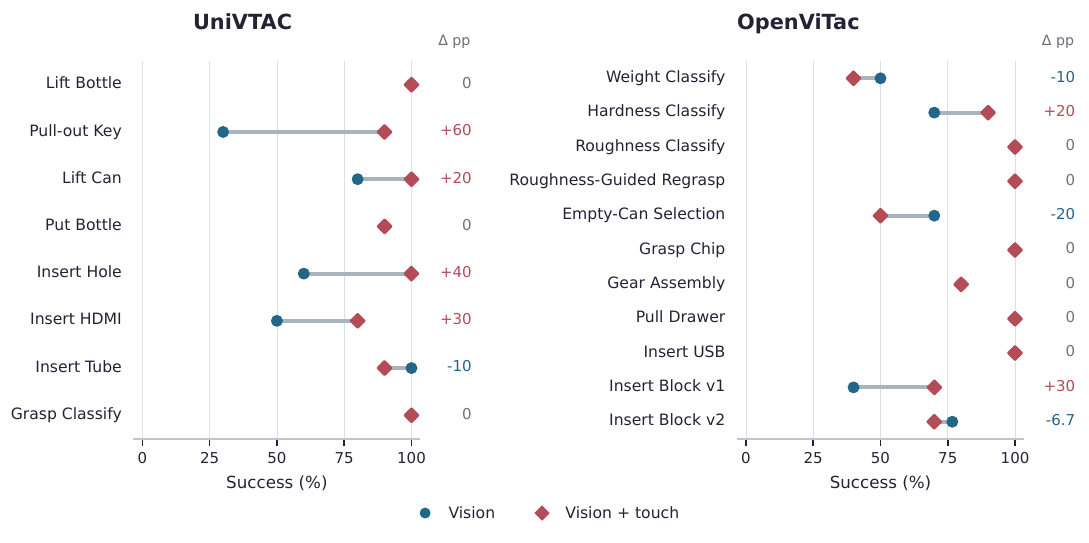}\caption{\textbf{Per-task changes with tactile input.} Circles denote \V, diamonds denote \VT, and the right-hand numbers show \VT\ minus \V\ in percentage points. Six tasks improve, nine are unchanged, and four decrease.}\label{fig:modality}\end{figure}

Adding tactile RGB input improves four UniVTAC tasks: Pull-out Key increases from 30\% to 90\%, Lift Can from 80\% to 100\%, Insert Hole from 60\% to 100\%, and Insert HDMI from 50\% to 80\%. On OpenViTac, Hardness Classify improves from 70\% to 90\%, and Insert Block v1 from 40\% to 70\%. The corresponding successful trajectories illustrate resistance detection, grasp stabilization, contact-guided alignment, and material-property judgment.

Vision-based tactile sensing encodes local contact in images~\citep{yuan2017gelsight} that Astra can process through its existing visual interface. In successful traces, changes in contact shape, symmetry, shear, and stability are associated with targeted adjustments: unloading a constrained object, changing the approach direction, testing retention, or limiting closure after contact. These behaviors demonstrate useful tactile interpretation through the same image interface used for ordinary scene perception. The successful interventions depend on the contact regime. Constrained insertion calls for realignment before renewed advancement; fragile grasping calls for stopping closure once support is established; material classification uses controlled changes in contact to obtain a more informative comparison with demonstrations.

The following eight successful trajectories illustrate representative manipulation behaviors, organized into contact-state inference, active probing, fine adjustment, and manipulation recovery. Each storyboard presents recorded observations and concise action summaries. Step labels are zero-based control-action indices, not model-request counts; selected intervals are nonuniform. Camera crops improve legibility, while tactile images retain their original content and aspect ratio. The analysis combines recorded action notes with visible execution. Action notes describe the policy assessment, and the recorded observations show the resulting motion and task outcome. Appendix~\ref{app:pairs} presents same-seed outcomes for the six tasks that improve with touch, connecting the task-level comparison to individual trajectories.

\subsubsection{Contact-State Inference}
A useful tactile policy must connect changing contact observations to an appropriate intervention. The first two cases illustrate responses to constrained motion and inference about a partially hidden insertion direction.
\begin{figure}[!htbp]\centering\includegraphics[width=\linewidth]{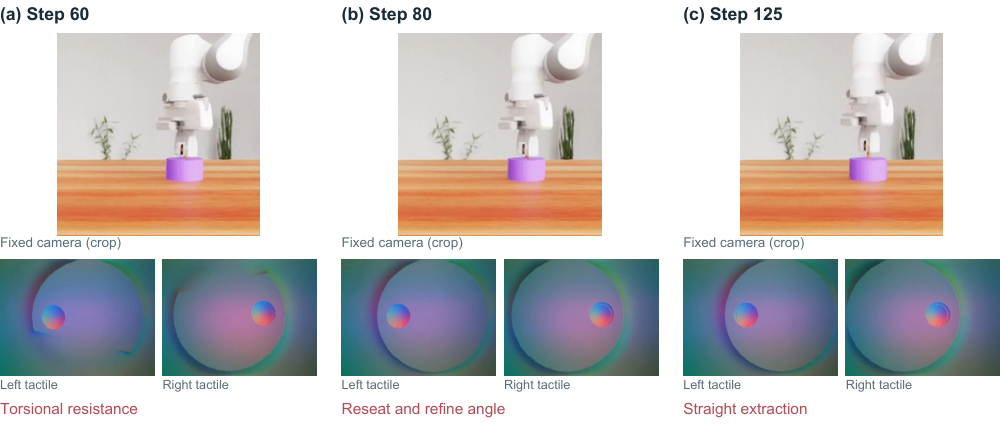}\caption{\textbf{Pull-out Key: resistance-aware extraction.} Seed 1000017, \VT, success. At step 60, the action note identifies torsional deformation and proposes unloading the twist. Continued axial resistance leads to reseating and finer angular adjustment at step 80; the later frame shows extraction. The sequence alternates extraction attempts with unloading and realignment as resistance persists. \record{UniVTAC_pull_out_key_pull_out_key_1000017_vision_tactile}.}\label{fig:key}\end{figure}

The informative feature of the Pull-out Key sequence (Figure~\ref{fig:key}) is the alternation between testing extraction and reducing the constraint that blocks it. Continued resistance changes the next action from another pull to unloading and angular refinement. This matters because larger displacement in the same direction would not resolve a poorly aligned or twisted contact. The later extraction provides a visible outcome for the revised strategy; the tactile interpretation is useful through the corrective action it supports.
\begin{figure}[!htbp]\centering\includegraphics[width=\linewidth]{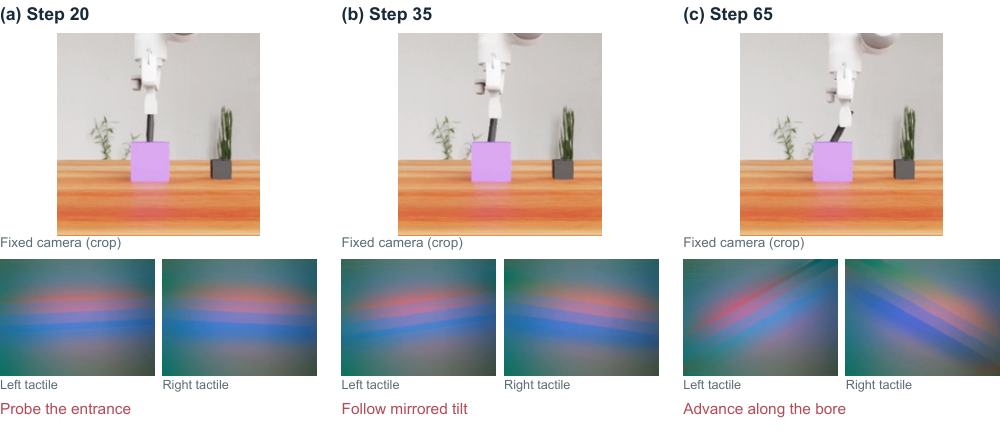}\caption{\textbf{Insert Hole: inferring the bore direction.} Seed 1000028, \VT, success. Astra interprets the opposing slopes of the tactile bands together with visible shaft deflection to infer a direction different from the reference trajectory. Astra pivots the shaft while maintaining entrance contact, then advances along the inferred bore. The case illustrates adaptation to contact geometry rather than literal replay of the demonstrated motion. \record{UniVTAC_insert_hole_insert_hole_1000028_vision_tactile}.}\label{fig:hole}\end{figure}

In Insert Hole (Figure~\ref{fig:hole}), the target direction is partly hidden by the surrounding geometry. The tactile pattern and visible shaft deflection provide a local reference for adjusting the approach. Astra tests its inferred direction through subsequent advancement, adapting the demonstrated procedure to the current contact geometry. The sequence shows how local observations support a new motion direction while preserving the overall task goal.

\subsubsection{Active Probing}
Astra sometimes acts to obtain a more discriminative observation before committing to a task decision. In Hardness Classify, a change in grip creates an informative response, but also requires recovery when the object begins to slip.
\begin{figure}[!htbp]\centering\includegraphics[width=\linewidth]{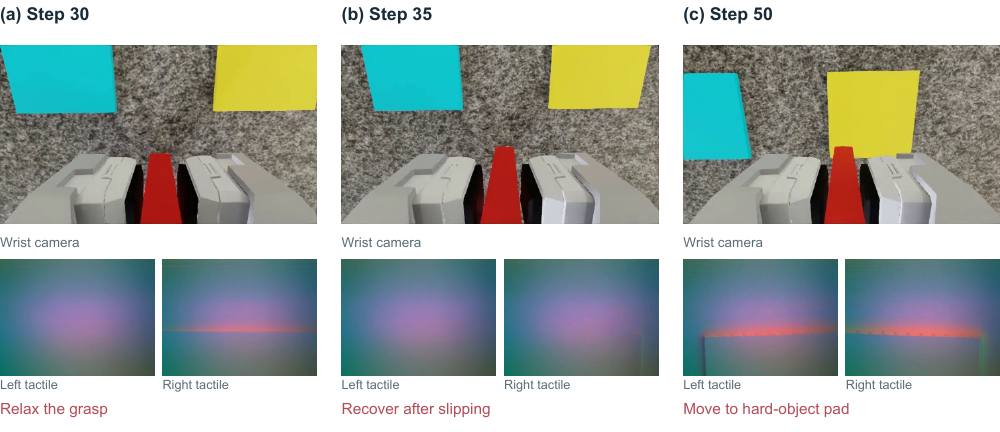}\caption{\textbf{Hardness Classify: probing while maintaining retention.} Seed 10006, \VT, success. Astra slightly relaxes the grasp after lifting, detects slipping, and re-establishes contact. The subsequent action note associates the sharply defined loaded edge with the hard-object demonstration. The probe changes the contact condition, and the subsequent correction preserves retention before classification. \record{OpenViTac_hardness_classify_hardness_classify_hard_10006_vision_tactile}.}\label{fig:hardness}\end{figure}

The Hardness Classify probe (Figure~\ref{fig:hardness}) is also a control intervention: relaxing the fingers changes both the observation and the security of the grasp. The subsequent recovery is therefore part of the probing behavior. Astra must maintain the object while obtaining evidence for its classification. The sequence connects a deliberately changed contact condition to a reference-based property judgment, instead of relying on a single unchanging tactile appearance.
\subsubsection{Fine Adjustment}
Near contact, small changes can determine whether a grasp succeeds without damaging the object. The Grasp Chip sequence separates first contact, centering, bilateral support, and the retention test.
\begin{figure}[!htbp]\centering\includegraphics[width=\linewidth]{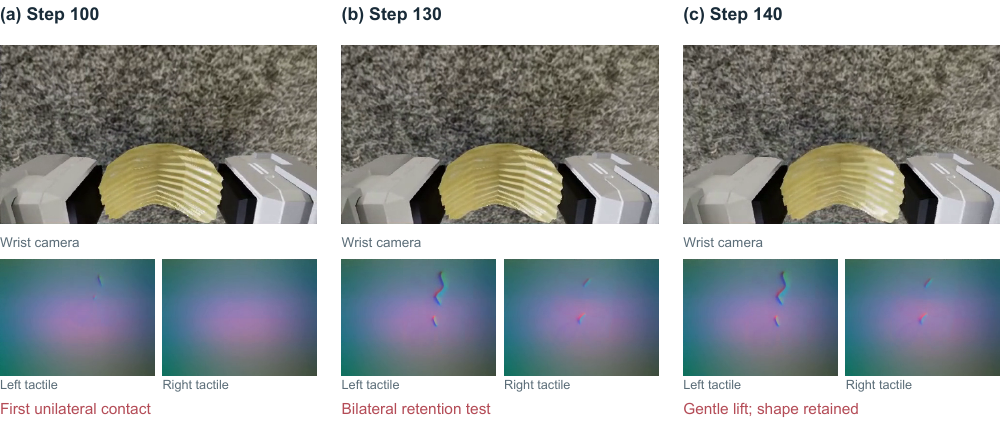}\caption{\textbf{Grasp Chip: cautious convergence on a fragile grasp.} Seed 10099, \VT, success. Unilateral contact triggers smaller closure increments and a commanded 0.5\,mm lateral correction at step 110, between panels (a) and (b). After bilateral contact develops, Astra avoids continued tightening and tests retention through a gentle lift. The episode later completes intact placement; the panels focus on the contact-establishment phase. \record{OpenViTac_grasp_fragile_chip_grasp_fragile_chip_10099_vision_tactile}.}\label{fig:chip}\end{figure}

The Grasp Chip sequence (Figure~\ref{fig:chip}) separates establishing contact from increasing grip strength. Initial contact on one side prompts a small positional correction, after which bilateral contact can be checked before lifting. This ordering is useful for a fragile object because continued closure is not the only way to improve retention. A gentle lift then tests whether the current support is sufficient. The successful outcome depends on coordinating these stages rather than issuing one aggressive closing command.

Successful correction requires identifying which part of an operation is blocked. Gear assembly separates center alignment from tooth engagement; USB insertion distinguishes hand displacement from actual connector progress.
\begin{figure}[!htbp]\centering\includegraphics[width=\linewidth]{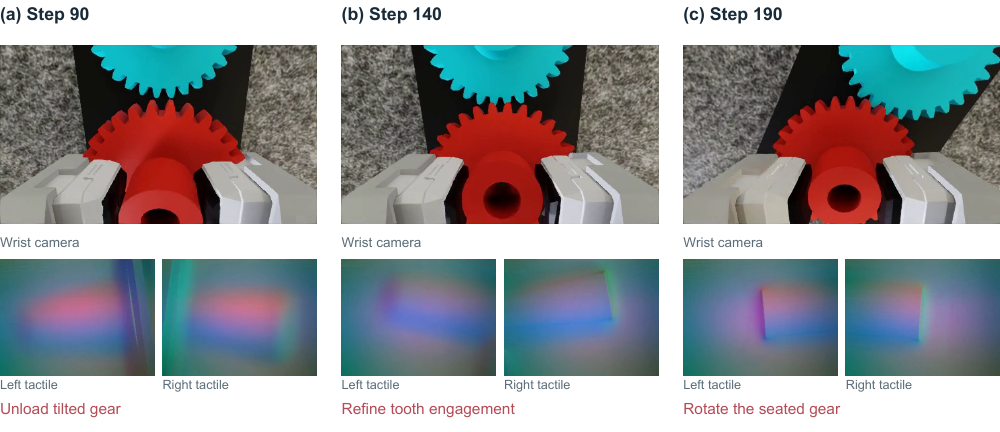}\caption{\textbf{Gear Assembly: resolving successive assembly constraints.} Seed 10034, \VT, success. When the red gear tilts, Astra unloads contact and adjusts its center. A later yaw adjustment addresses possible tooth interference before seating and rotation. The correction strategy changes from centering to tooth engagement as the observed constraint changes. \record{OpenViTac_turn_gear_pair_turn_gear_pair_10034_vision_tactile}.}\label{fig:gear}\end{figure}

The gear in Figure~\ref{fig:gear} presents successive constraints at different stages of assembly. Recentering can address a tilted approach, but it does not by itself establish a compatible tooth configuration. The later angular change reflects this distinction. The observed behavior is therefore more specific than a general instruction to move closer to the target: the policy revises the type of correction as the operation progresses from gross alignment toward engagement and seating.
\begin{figure}[!htbp]\centering\includegraphics[width=\linewidth]{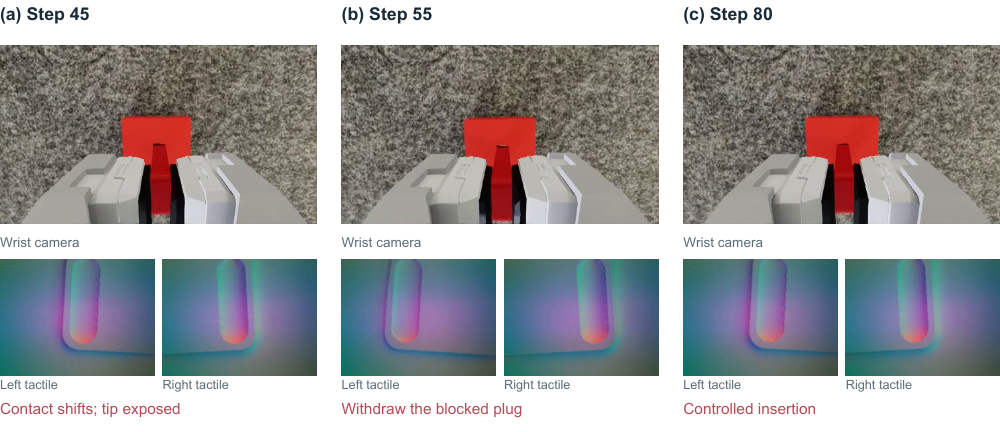}\caption{\textbf{Insert USB: verifying object progress.} Seed 10066, \VT, success. The connector remains exposed while the tactile contacts shift, and the action note identifies finger slip along a blocked plug. Astra withdraws, recenters, and issues a submillimeter correction before insertion progresses. Hand motion is not treated as sufficient evidence that the connector has entered the socket. \record{OpenViTac_insert_USB_insert_USB_10066_vision_tactile}.}\label{fig:usb}\end{figure}

The contrast between hand motion and object progress is central to the USB case (Figure~\ref{fig:usb}). A moving gripper can change the tactile contact even when the plug remains blocked. The exposed connector and shifting finger contacts help distinguish these possibilities. Withdrawing and recentering changes the contact configuration before another advance. This is a useful form of progress checking because issuing a displacement command alone does not establish that the intended insertion has occurred.

Grasp preparation and local correction can both influence insertion. The successful Insert Block v1 case combines an initial orientation adjustment with a small retreat and recentering near the opening.
\begin{figure}[!htbp]\centering\includegraphics[width=\linewidth]{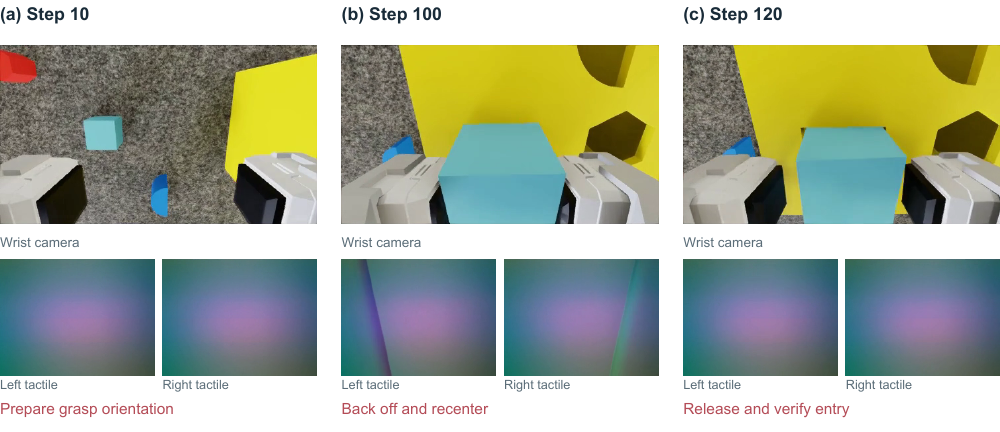}\caption{\textbf{Insert Block v1: a local correction after rim contact.} Seed 10036, cube, \VT, success. Astra prepares the hand orientation before grasping, transports the cube, and then responds to possible rim contact with a small retreat and a commanded 2\,mm centering adjustment. The final frame shows the release-and-verification phase. The local retreat creates clearance for recentering before renewed insertion. \record{OpenViTac_insert_block_v1_insert_block_v1_cube_layout_014_10036_vision_tactile}.}\label{fig:block1}\end{figure}

In Insert Block v1 (Figure~\ref{fig:block1}), the initial grasp is part of the alignment problem: the chosen hand orientation influences how the cube is presented to the opening. Near the rim, the policy uses a different scale of correction, combining a small retreat with millimeter-scale recentering. The retreat creates an opportunity to change the approach before renewed entry. The episode links preparation at a distance with local adjustment near contact, while the final release checks the resulting placement.
\subsubsection{Manipulation Recovery}
A release command does not establish that an insertion has succeeded. The next case checks the resulting object state and reopens the manipulation sequence when the part remains on the rim.
\begin{figure}[!htbp]\centering\includegraphics[width=\linewidth]{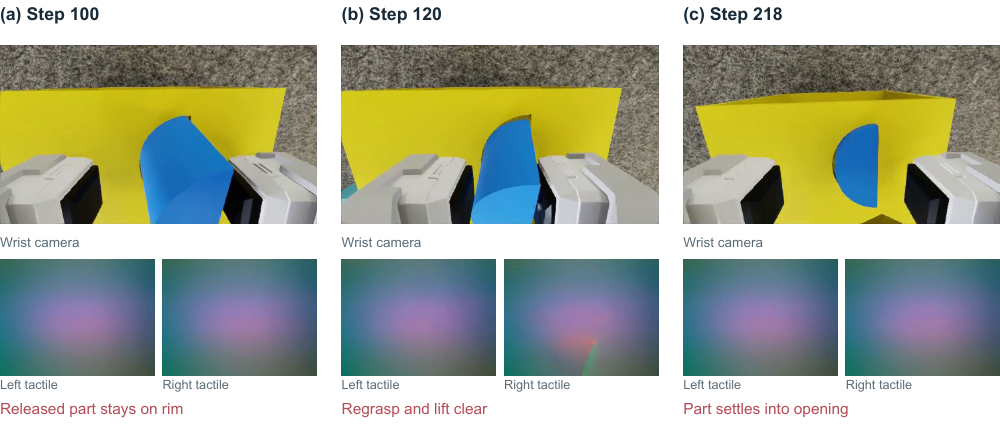}\caption{\textbf{Insert Block v2: recovering from unsuccessful placement.} Seed 10019, half-cylinder, \VT, success. After release, the part remains at the opening. Astra regrasps, lifts, and makes further alignment attempts until the part settles into the opening. The behavior demonstrates outcome verification and recovery from an incomplete placement. \record{OpenViTac_insert_block_v2_half_cylinder_insert_block_v2_half_cylinder_layout_132_10019_vision_tactile}.}\label{fig:block2}\end{figure}

Recovery in Insert Block v2 (Figure~\ref{fig:block2}) begins with a discrepancy between the intended and observed result. The release action has finished, but the object has not reached the required state. Astra consequently treats the task as unfinished and initiates another grasp-and-alignment sequence. This distinction between command completion and task completion is essential in contact-rich manipulation, where a plausible action sequence can leave an object balanced on an edge or obstructed at an entrance.

\subsection{Visual Compensation for Missing Touch}\label{sec:vision}
The gain associated with tactile input differs substantially between benchmarks: 17.5 percentage points on UniVTAC and 1.2 points on OpenViTac. Their camera configurations provide one possible explanation. UniVTAC restricts some visual cues in its tactile evaluation setting: six of its eight tasks use only the fixed camera, while Lift Can and Insert Tube also use a wrist camera. OpenViTac provides both fixed and wrist cameras across its tasks, yielding richer visual observations. These differences may reduce the marginal value of touch for a model that can identify subtle visual structure.

\begin{figure}[!htbp]\centering\includegraphics[width=\linewidth]{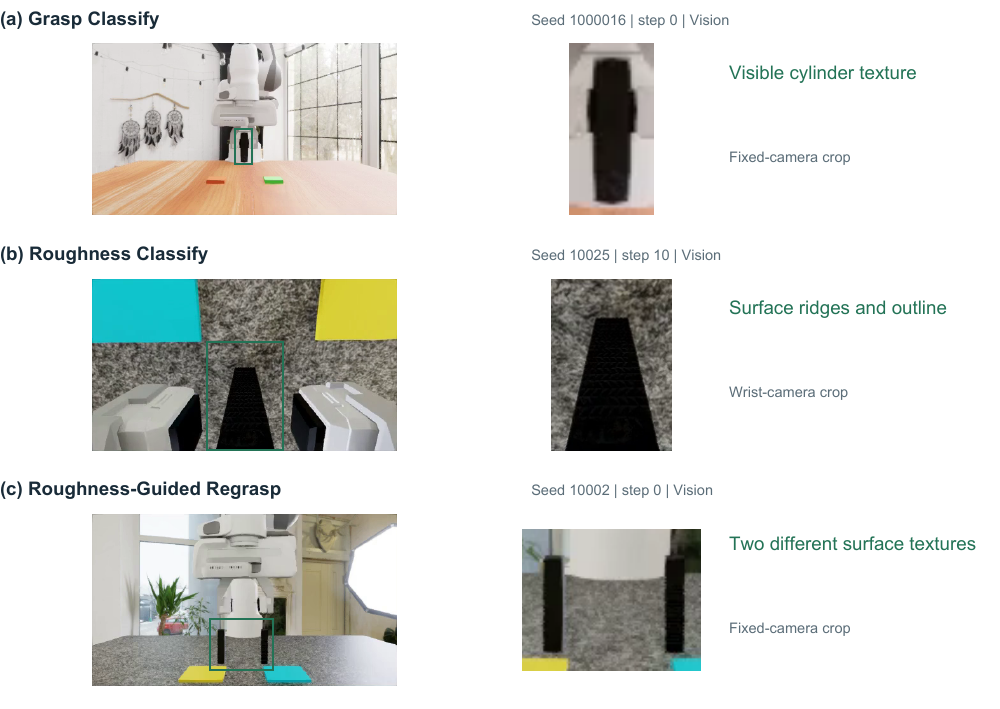}\caption{\textbf{Visible cues in nominally tactile tasks.} Three \V\ episodes from tasks with 100\% sampled success. Left: a recorded camera view; center: enlargement of the green-marked region; right: the visible cue. Grasp Classify shows cylinder texture, Roughness Classify exposes surface ridges and contour structure, and Roughness-Guided Regrasp contains visually different surfaces even in the fixed-camera view. Enlargements preserve the recorded pixels. Rollouts: \href{https://tactalent.github.io/gallery.html?lang=en\&episode=UniVTAC_grasp_classify_grasp_classify_1000016_vision}{Grasp Classify}, \href{https://tactalent.github.io/gallery.html?lang=en\&episode=OpenViTac_roughness_classify_roughness_classify_rough_10025_vision}{Roughness Classify}, and \href{https://tactalent.github.io/gallery.html?lang=en\&episode=OpenViTac_roughness_regrasp_roughness_regrasp_left_10002_vision}{Roughness-Guided Regrasp}.}\label{fig:visual}\end{figure}

In all three tasks in Figure~\ref{fig:visual}, Astra identifies visible texture or ridges in its action notes and uses these cues to complete the required classification or selection without tactile images. This level of pure-vision discrimination goes beyond what the reported specialist results demonstrate. Its sensitivity exposes subtle details that human observers can easily overlook, showing why visually inconspicuous cues still matter when designing tasks intended to require touch.

The three cases expose different routes to visual discrimination. Grasp Classify provides texture on the cylinder surface; Roughness Classify adds a close view of ridges and object contour; Roughness-Guided Regrasp presents distinguishable surfaces in the fixed-camera scene. Such cues can support the required category or selection decision even without directly measuring the physical property through touch. Their usefulness depends on the correspondence between visible appearance and the task categories, as well as the model's ability to notice it.

The visual evidence is not limited to static texture. In three inspected Weight Classify pairs where \V\ succeeds and \VT\ fails, the vision-only policy relaxes the grasp and uses visible slip to reach the correct category. The tactile runs instead associate weak shear or a contact pattern resembling the light-object demonstration with low weight and misclassify the heavy object. Empty-Can Selection shows a related ambiguity: subtle contact changes can be misread, while repeated comparisons leave insufficient actions for transfer. These observations connect the visual-compensation result to concrete failures in evidence interpretation and probe selection, examined in Section~\ref{sec:limits}.

\subsection{Scaling Potential of Vision-Based Touch}\label{sec:scaling}
The preceding results identify an important property of vision-based tactile sensing: it expresses contact through an interface that a general multimodal model already processes. GelSight-style sensors convert local deformation into images~\citep{yuan2017gelsight}; the model can relate these images to object appearance, robot state, preceding motion, and demonstrated outcomes. In our experiments, that relationship supports decisions about constrained motion, retention, and alignment without training a new tactile policy for each task.

This observation complements work on reusable tactile representations. VITaL, Sparsh, AnyTouch, and T3 learn representations that transfer across tactile settings~\citep{george2024vital,higuera2024sparsh,feng2025anytouch,zhao2025t3}; FTP-1 and SimpleTouch connect contact-sensitive representations to action generation~\citep{yuan2026ftp,yang2026simpletouch}. Astra exposes a further route: a single general visual interface can interpret both scene images and contact images through demonstration-conditioned inference. The successful behaviors in Section~\ref{sec:tactile} illustrate how temporal context connects contact observations to corrective actions. A contact image becomes actionable when the model connects it to what the fingers did and what happened to the object.

Joint visual--tactile pretraining is a natural next step. Aligned scene images, tactile images, actions, and outcomes could make this correspondence part of general multimodal learning rather than a relationship supplied only at deployment. The resulting research question concerns how contact understanding improves with data diversity, model capacity, and interaction experience. Our results motivate this direction through the capability already accessible from tactile RGB inputs. For sensor development, they favor representations that preserve interpretable contact structure and support comparison across sensors, grasps, and motion conditions.

\subsection{Emerging Benchmark Saturation}\label{sec:saturation}
With tactile input, Astra reaches at least 90\% success on thirteen of nineteen tasks, including perfect sampled success on nine. UniVTAC's mean is 93.8\%, while several OpenViTac classification and transfer tasks also reach 100\%. These ceilings limit how well the corresponding tasks distinguish stronger policies. They also shift attention from average completion toward the quality of contact decisions, recovery, and generalization.

Visual compensation adds a separate benchmark-design requirement. A tactile task should isolate information that the chosen visual interface does not already reveal. In the current suites, texture, contour, or motion can correlate with the required physical judgment. Astra's results demonstrate that such cues become useful to a sufficiently capable visual model, even when they are not salient to a human observer. UniVTAC's restricted camera access and OpenViTac's richer sim-and-real observation setting produce complementary tests of this issue~\citep{chen2026univtac,wu2026openvitac}. Both contact fidelity and control over alternative visual evidence matter when interpreting a tactile-policy result.

More discriminative evaluations should include visually indistinguishable objects with different mass, friction, stiffness, or internal constraints; tighter insertion tolerances; longer contact sequences; and recovery after disturbed grasps or failed engagement. Varying appearance independently of physical properties would test whether a policy infers contact rather than memorizes a visual correlate. Evaluating the same judgments across sensors and real robots would further separate robust tactile understanding from a particular rendering or contact configuration. The broader benchmark goal is reliable physical inference and correction, not simply a larger set of scenes with the same available cues.

\FloatBarrier
\section{Remaining Limitations of Astra}\label{sec:limits}
Tactile input does not consistently improve performance: Weight Classify, Empty-Can Selection, Insert Tube, and Insert Block v2 decrease in the sampled results. Three recurring limitations emerge from the trajectories: limited sensitivity to tangential motion and force, difficulty with fine adjustments, and active probing that fails to converge. Weak tactile cues can be misread, a plausible correction can fail to change the object configuration as intended, and repeated observations can leave too little time to finish the manipulation.
\subsection{Limited Sensitivity to Tangential Motion and Force}
\begin{figure}[!htbp]\centering\includegraphics[width=\linewidth]{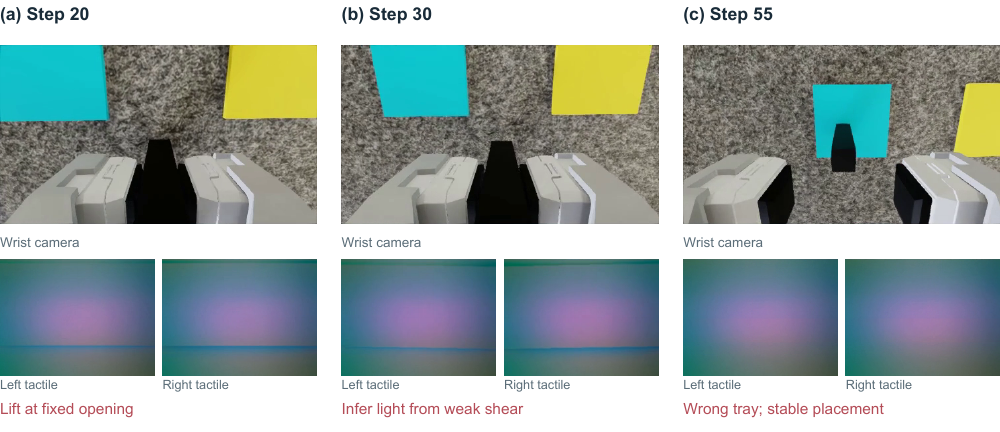}\caption{\textbf{Limited sensitivity to tangential motion and load.} Weight Classify, seed 10038, \VT, failure. Astra interprets small tactile changes during a lift as evidence of a light object and places the heavy block on the incorrect blue tray. The image pattern also depends on grasp position, aperture, and motion; a small shear response is therefore not an unambiguous weight measurement. \record{OpenViTac_weight_classify_weight_classify_heavy_10038_vision_tactile}.}\label{fig:weight-fail}\end{figure}

The ambiguity in Weight Classify (Figure~\ref{fig:weight-fail}) lies in relating the image change to the quantity being inferred. During a lift, the observed contact depends on how the object is supported and whether it moves relative to the fingers. A small change may therefore be insufficient to distinguish low load from a contact configuration that transmits little visible shear. In this episode, Astra commits to the light-object interpretation without resolving that ambiguity, showing a gap between recognizing tactile variation and reliably interpreting tangential loading.
\subsection{Fine Adjustments Remain Challenging}\label{sec:contact-diagnosis}
\begin{figure}[!htbp]\centering\includegraphics[width=\linewidth]{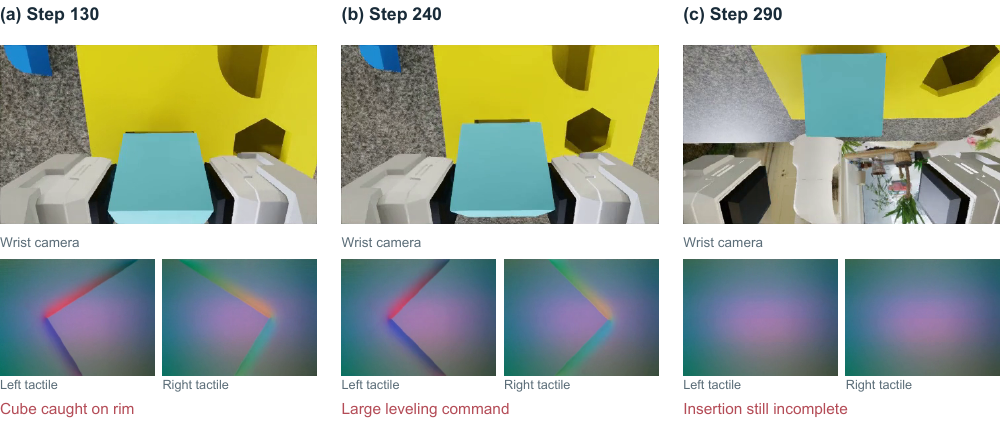}\caption{\textbf{Recognizing misalignment does not ensure effective correction.} Insert Block v1, seed 10016, \VT, failure. Repeated rim contact prompts regrasping and leveling. From step 240 onward, the cumulative commanded pitch increments total approximately $51.6^\circ$, with compensatory translation, yet insertion remains incomplete at the action limit. \record{OpenViTac_insert_block_v1_insert_block_v1_cube_layout_132_10016_vision_tactile}.}\label{fig:align-fail}\end{figure}

The Insert Block v1 failure (Figure~\ref{fig:align-fail}) also separates diagnosing an error from executing a successful correction. A large accumulated rotation command does not specify the object's actual rotation under contact. Translation, rotation, and the location of support must remain compatible for the part to clear the rim and enter. The repeated attempts show that identifying the need to level the cube is insufficient when the resulting motion does not produce the required object configuration.

\begin{figure}[!htbp]\centering\includegraphics[width=\linewidth]{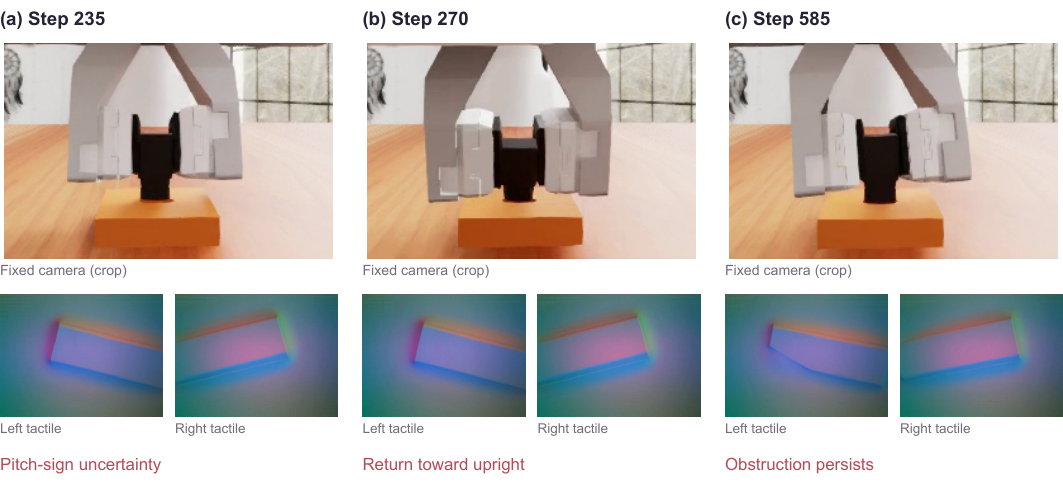}\caption{\textbf{Recognizing obstruction without resolving the correction direction.} Insert HDMI, seed 1000062, \VT, failure. Fixed-camera crops and paired tactile images show three stages of the same unsuccessful episode. The recorded notes identify pitch-sign uncertainty at action 235, a return toward an upright interpretation at 270, and persistent obstruction at 585. The changing contact patterns do not lead to successful insertion. \record{UniVTAC_insert_HDMI_insert_HDMI_1000062_vision_tactile}.}\label{fig:diagnosis-fail}\end{figure}

Detecting an obstruction does not by itself identify the correction needed to remove it. The failed Insert HDMI episode in Figure~\ref{fig:diagnosis-fail} makes this distinction explicit. At action 235, Astra states that the tactile image-axis sign is uncertain and tests the opposite pitch direction. At action 270, it returns toward an upright interpretation after the alternate pitch still produces slip. At action 585, the action note again identifies obstruction rather than seating. The episode ends without success after 600 actions and 120 model decisions.

The contact evidence triggers plausible adjustments, but the sequence does not converge on a reliable relationship between the observed deformation and the corrective motion. Depth, pitch, and yaw are revised across attempts while the plug remains near the socket mouth. A new contact pattern can therefore prompt another hypothesis without isolating why the previous attempt failed. Effective diagnosis requires probes that discriminate between competing explanations and subsequent decisions that preserve what those probes established. Together with the Insert Block v1 example, this sequence distinguishes detecting a failed action, identifying its physical cause, and producing an effective correction.

\subsection{Active Probing May Fail to Converge}\label{sec:probing-limits}
Active probing must reduce uncertainty while preserving enough actions to complete the task. Repeated observations are useful only when they lead to a timely decision about how to proceed.

The Weight Classify comparison in Section~\ref{sec:vision} illustrates how an added modality changes the evidence prioritized by the policy. Comparing contacts under consistent probing conditions is essential for distinguishing a material or load difference from a change in support or grasp geometry.

Empty-Can Selection reveals a related efficiency problem. Repeated measurements can yield a plausible eventual judgment while leaving insufficient opportunity to finish the task. The issue is not only perceptual accuracy, but also whether a probe resolves uncertainty at a reasonable cost in remaining actions.
\begin{figure}[!htbp]\centering\includegraphics[width=\linewidth]{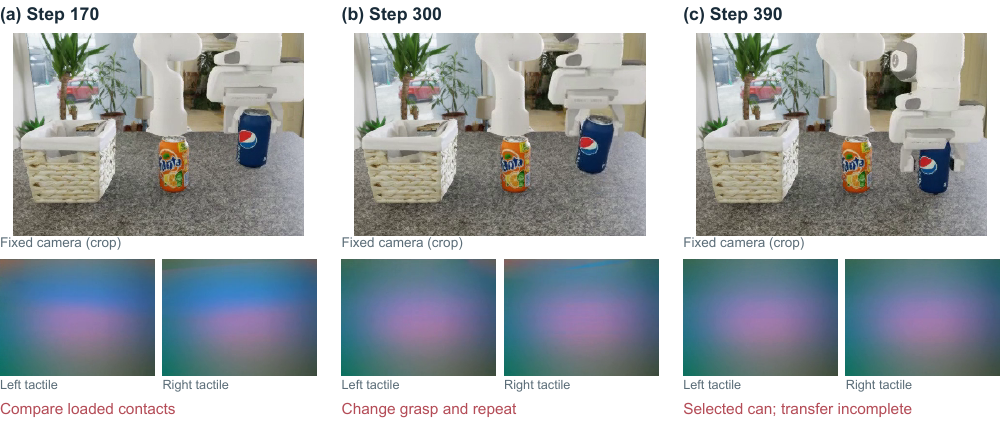}\caption{\textbf{Repeated probing without timely convergence.} Empty-Can Selection, seed 10034, \VT, failure. Astra compares load responses, changes the grasp after unstable contact, and repeats the comparison. It selects the correct blue can at step 370 and starts the final grasp-and-lift sequence at step 390, but exhausts the 400-action budget before transfer. Information gathering must be coordinated with a stopping rule and the remaining manipulation. \record{OpenViTac_can_empty_select_can_empty_select_pepsi_10034_vision_tactile}.}\label{fig:probe-fail}\end{figure}

In the Empty-Can Selection failure (Figure~\ref{fig:probe-fail}), the eventual category judgment and the completed manipulation are separate outcomes. A probe is useful only if the information it produces can still guide the remaining grasp and transfer. Repeatedly changing the contact conditions also makes successive observations harder to compare directly. The correct selection is recorded only at action 370, leaving 30 of 400 actions for the remaining grasp, transport, and placement. This trajectory exposes a stage-transition problem: gathering more evidence must be balanced against the manipulation still required after the classification decision. Consistent probing conditions and a stopping rule should preserve enough opportunity to finish those later stages.

These failures expose a gap between locally plausible actions and reliable task completion. Astra must relate contact cues to the underlying physical constraint, turn that diagnosis into effective corrective motion, and stop probing in time to complete the remaining manipulation.

\FloatBarrier
\section{Inference Cost and Latency}\label{sec:cost}
The two evaluation phases consume a combined 335.97 million tokens and CNY\,7,505.05 in logged provider cost. The \VT\ phase uses approximately 59\% more tokens and incurs 42\% more cost than \V. Cached inputs account for more than 80\% of input tokens in both settings, reflecting the repeated context supplied during closed-loop control.
\begin{table}[!htbp]\centering
\caption{\textbf{Evaluation-phase token usage and cost.} \V: vision; \VT: vision and tactile RGB. Phase totals include billed errors, retries, and reruns, but exclude demonstration preparation and later supplementary experiments. Cached input is included in input tokens; reasoning tokens are included in output tokens. Costs sum the provider's logged estimates.}\label{tab:cost}
\vspace{4pt}
\small\renewcommand{\arraystretch}{1.2}
\begin{tabularx}{\linewidth}{@{}l*{5}{C}@{}}
\toprule
Input & Input tokens & Cached input & Output tokens & Total tokens & CNY\\
\midrule
\V &127,250,755&104,837,109&2,283,555&129,534,310&3,102.25\\
\VT &203,911,092&170,693,217&2,521,218&206,432,310&4,402.80\\
\midrule
Total &331,161,847&275,530,326&4,804,773&335,966,620&7,505.05\\
\bottomrule
\end{tabularx}
\end{table}

The median complete-response latency is 15.4\,s for \V\ and 14.0\,s for \VT; the respective 90th percentiles are 46.3\,s and 44.0\,s. Both distributions are right-skewed: responses longer than 120\,s account for 0.76\% and 0.47\%, respectively. Latency measures the complete API response for a model request, including provider and gateway waiting time and excluding simulation execution. Each response can generate several control actions. The two modes exhibit similar latency distributions across their respective request mixes and service conditions.

\begin{figure}[!htbp]\centering\includegraphics[width=\linewidth]{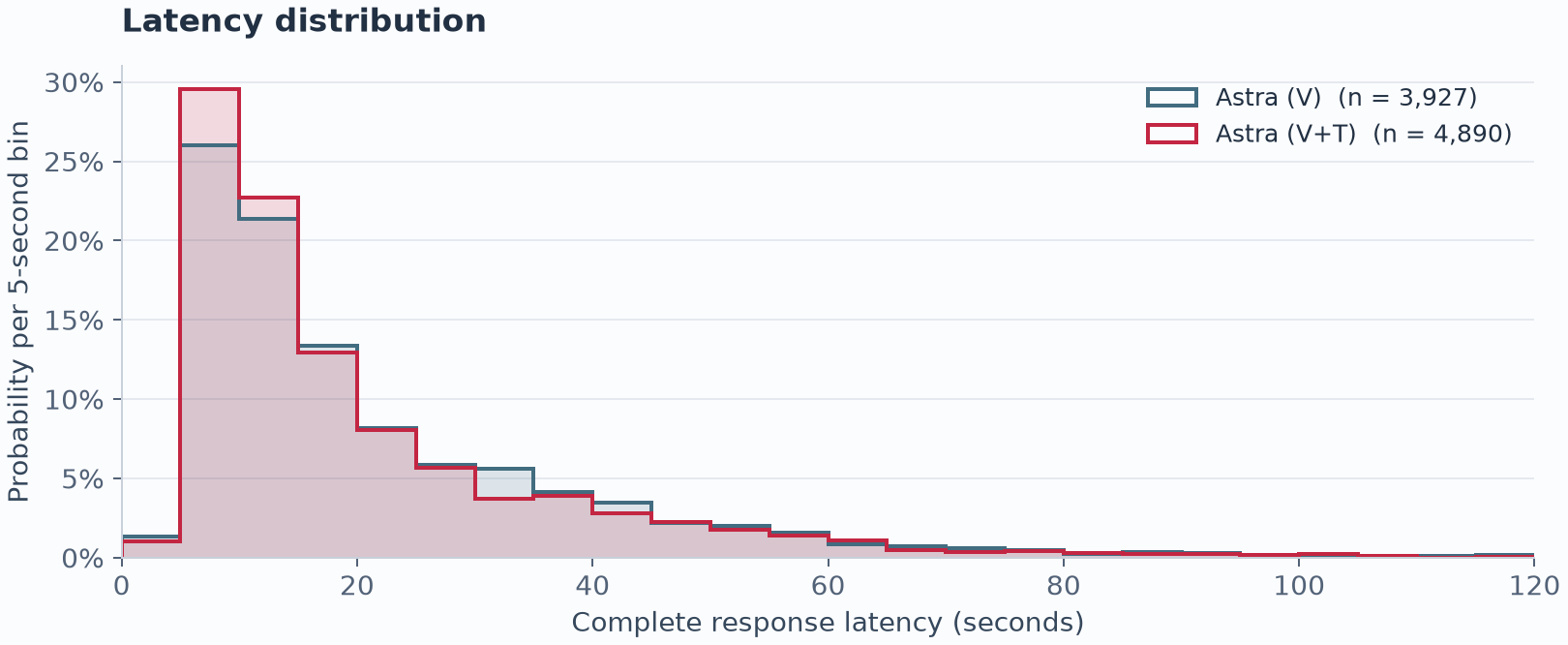}\caption{\textbf{Distribution of API response latency.} Five-second bins, normalized by the total number of valid responses in each setting: 3,927 for \V\ and 4,890 for \VT. The plotted range ends at 120\,s; the omitted tails remain in the normalization. Failed requests and responses without output tokens are excluded from latency statistics.}\label{fig:latency}\end{figure}

\begin{table}[!htbp]\centering
\caption{\textbf{Complete API response latency in seconds.} \V: vision; \VT: vision and tactile RGB. Each row summarizes valid, deduplicated responses with positive output tokens. P90, P95, and P99 denote the 90th, 95th, and 99th percentiles. Mean and quantiles include the full distribution, including responses beyond the displayed histogram range.}\label{tab:latency}
\vspace{4pt}
{\small\setlength{\tabcolsep}{4pt}\renewcommand{\arraystretch}{1.15}
\begin{tabularx}{\linewidth}{@{}l*{6}{C}@{}}\toprule
Input & Requests & Mean & Median & P90 & P95 & P99\\\midrule
\V &3,927&22.8&15.4&46.3&59.1&110.7\\
\VT &4,890&21.1&14.0&44.0&56.8&96.6\\\bottomrule
\end{tabularx}}
\end{table}

The response tails summarized in Table~\ref{tab:latency} delay subsequent decisions in closed-loop operation, since each decision depends on the preceding execution and response. Although each response supplies several control actions, the next observation-dependent correction must wait for another model response. This dependency makes response latency a key factor in end-to-end deployment.

\section{Limitations of This Study}\label{sec:study-limitations}
All experiments are conducted in simulation; robustness on physical robots remains to be validated under sensor noise and drift, calibration error, actual contact dynamics, and execution delays. Cost constraints limit most tasks to ten episodes per input setting instead of the full 100-episode evaluation, so a single outcome changes the task success rate by ten percentage points. Larger samples, multiple sampling seeds, and repeated executions within the same scenes are needed to quantify variation from scene selection, model decisions, and execution. The experiments also cover only two benchmarks and use task-specific successful demonstrations, with corresponding categories or shapes represented where applicable. Generalization to unfamiliar objects, new tasks, different tactile sensors and robot configurations, and longer multistage manipulation requires systematic evaluation.

\section{Conclusion}\label{sec:conclusion}
GPT-6 Astra directly interprets vision-based tactile images and uses their contact cues for manipulation, achieving 93.8\% on UniVTAC and 81.8\% on OpenViTac and exceeding the reported specialist baselines in our comparison. Tactile images support contact-state inference, cautious grasp formation, alignment, and recovery; fine-grained vision also reveals cues that partially compensate for missing touch. These findings challenge the view that tactile manipulation is intrinsically reserved for specialist policies. The next step is to develop the capability and its evaluation together: joint visual--tactile pretraining can connect contact understanding to foundation-model scaling, while harder tasks with visually hidden physical properties, tighter tolerances, and repeated evaluation on real robots can establish reliable physical competence. The remaining errors in tangential-load interpretation, contact diagnosis, probe selection, and corrective control provide concrete targets for improving policy reliability and the execution harness.

\FloatBarrier
% A slightly smaller bibliography keeps complete references together.
\setlength{\bibsep}{1.5pt}
\renewcommand{\bibfont}{\fontsize{9.8}{11.8}\selectfont}

\clearpage
\appendix
\section{Task-Level In-Context Configuration}\label{app:configuration}
Table~\ref{tab:configuration} summarizes the archived demonstration configuration. A shot is a distinct demonstration trajectory. Each segment contains $H$ consecutive actions with the associated starting and resulting observations; segments are distributed across the demonstration rather than selected online by the policy. The same trajectories, segment starts, and actions are used in \V\ and \VT. Only \VT\ additionally includes the two tactile RGB views in demonstrations and online observations.

\begin{table}[!htbp]\centering
\caption{\textbf{Demonstrations, segment lengths, and camera inputs.} ``Shots'' counts distinct demonstration trajectories; ``Segments'' is the number selected per demonstration. $H$ is the number of actions per segment. Total actions counts demonstration actions in context, not an evaluation budget. The camera column lists official scene views; the visuo-tactile setting additionally receives both tactile RGB views. Insert Block v2 supplies all three shape demonstrations to each variant.}\label{tab:configuration}
\vspace{5pt}
{\small\setlength{\tabcolsep}{4pt}\renewcommand{\arraystretch}{1.3}\input{table-configuration.tex}}
\end{table}

\paragraph{Temporal Selection of Demonstration Blocks.}
For a source trajectory with $N$ observations, each selected start $s$ defines $H$ consecutive actions and the two observations at $s$ and $s+H$. Our preparation calls the released RoboICL deterministic selector~\citep{liu2026roboicl}: temporal targets distribute non-overlapping blocks across the trajectory, including its initial and terminal windows, and invalid action windows are excluded. The same selector is used with twelve blocks for one-shot context and five blocks per trajectory for multi-shot context. It selects blocks offline; online actions do not change their starts. For example, the two Weight Classify demonstrations use starts $\{0,33,66,99,132\}$ and $\{0,35,69,104,139\}$ with $H=5$. The corresponding images are the block endpoints, rather than every intermediate action frame.

\paragraph{Execution and Observation Timing.}
Demonstrations and generated commands use a 60\,Hz control cadence, with two 120\,Hz simulator substeps per command. Demonstration motion increments and gripper targets are reconstructed from recorded states under the benchmark's position-control interface. Both input settings use this execution protocol.

\clearpage
\section{Paired Outcomes on Touch-Improved Tasks}\label{app:pairs}
Paired episodes share the evaluation scene and random seed. Panels show each trajectory's final recorded frame with its own action index. Percentages report task-level success rates. For visual clarity, the figures display one scene-camera view; \VT\ episodes additionally use paired tactile images during control.
\begin{figure}[!htbp]\centering\includegraphics[width=\linewidth]{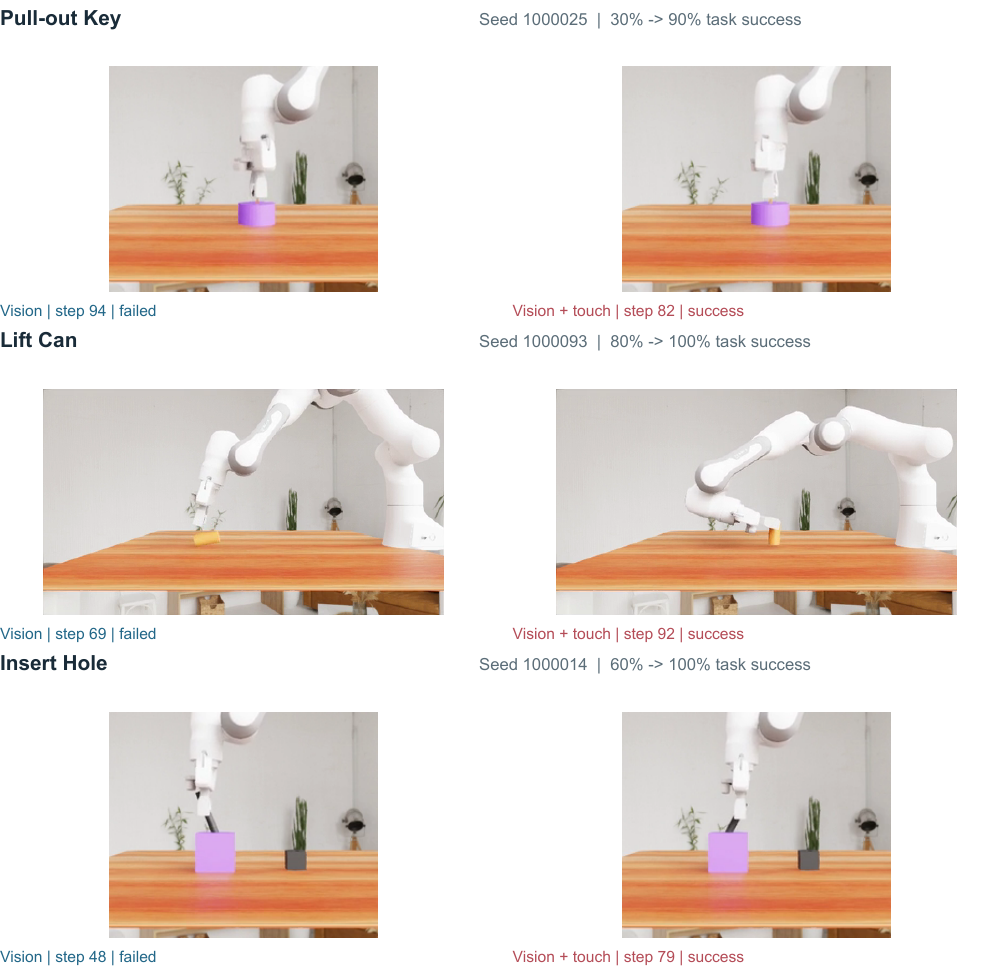}\caption{\textbf{Paired outcomes, part I.} Pull-out Key (seed 1000025), Lift Can (1000093), and Insert Hole (1000014). The \V\ episodes fail and the corresponding \VT\ episodes succeed. These endpoint views complement the successful contact-inference sequences in Figures~\ref{fig:key} and~\ref{fig:hole}.}\label{fig:pairs1}\end{figure}

\textbf{Pull-out Key.} The vision-only run repeatedly turns at fixed depth. With touch, growing edge deformation prompts torsional unloading; extraction begins after the contact returns to a round shape. The contact change guides the transition to axial withdrawal.

\textbf{Lift Can.} The vision-only run follows a tipping arc and later reports possible slip. The visuo-tactile run checks bilateral retention before lifting, then rotates the object upright and lowers it without releasing.

\textbf{Insert Hole.} Vision-only control probes a positive-X slant. Oppositely tilted tactile bands instead guide a negative-X pivot while the tip remains at the entrance, followed by insertion along the inferred bore. Touch helps select the hidden direction before advancement.

\clearpage
\begin{figure}[!htbp]\centering\includegraphics[width=\linewidth]{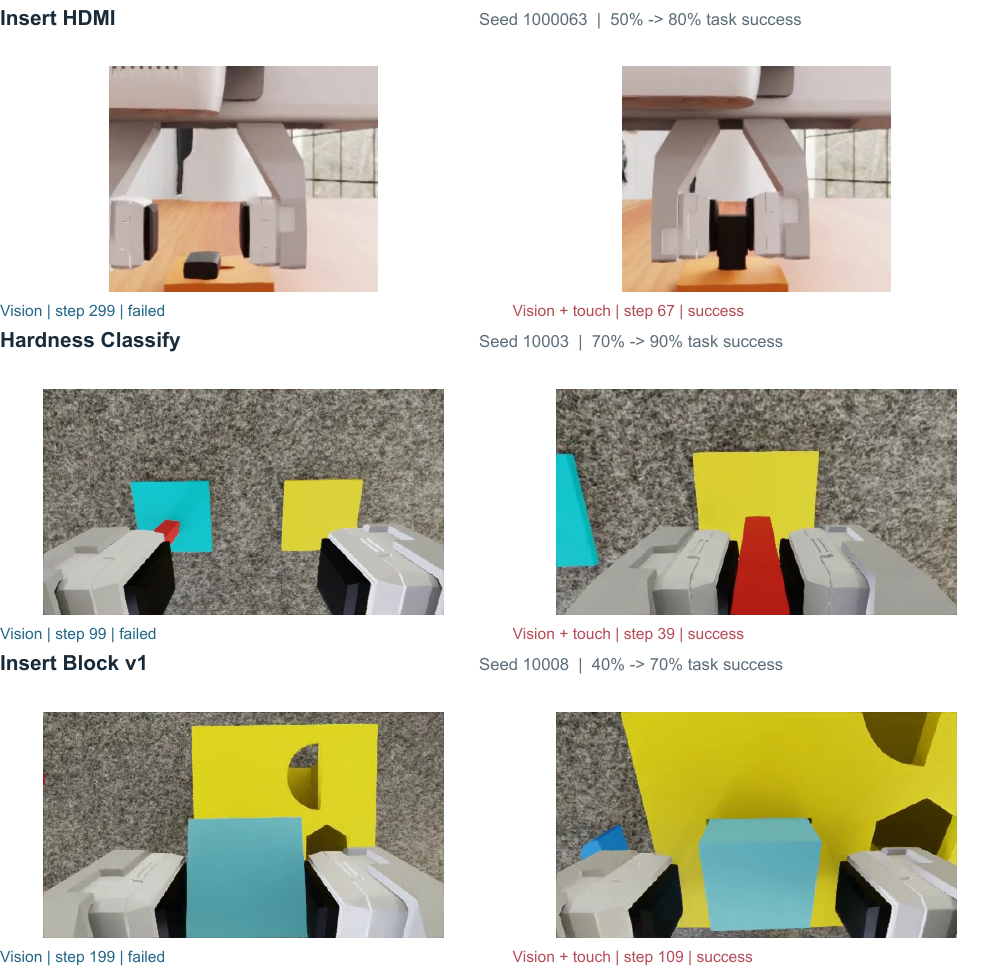}\caption{\textbf{Paired outcomes, part II.} Insert HDMI (seed 1000063), Hardness Classify (10003), and Insert Block v1 (10008). The \VT\ episodes succeed where the same-seed \V\ episodes fail. Hardness Classify requires the correct category-specific destination; the insertion tasks require the part to enter its matching opening. Complete action sequences and input views remain available in the online archive.}\label{fig:pairs2}\end{figure}

\textbf{Insert HDMI.} The vision-only run ends during another attempt to regrasp the displaced connector. With touch, shifting contact without plug motion is interpreted as rim contact, prompting unloading and fore-aft corrections. Advancement resumes once the plug descends with stable tactile contact. This sequence connects contact interpretation to correction of the object-to-socket alignment.

\textbf{Hardness Classify.} Vision-only control labels the hard block as compliant, places it on the blue plate, and continues observing the stable placement. The visuo-tactile run probes the grasp, associates a sharply bounded indentation with the hard category, and moves to the yellow plate. The pair distinguishes secure object transport from the material judgment that determines the required destination.

\textbf{Insert Block v1.} Both runs reach the square opening. In the vision-only trace, the cube remains on the rim after release, leading to repeated regrasping and centering attempts. The visuo-tactile run maintains retention and checks alignment before a shallow descent and gradual release complete insertion. The outcome depends on maintaining alignment from transport through release.

\end{document}

%% file: table-univtac.tex
\begin{tabularx}{\linewidth}{@{}l*{9}{C}@{}}
\toprule
\textbf{Model} & \makecell[c]{Lift\\Bottle} & \makecell[c]{Pull-out\\Key} & \makecell[c]{Lift\\Can} & \makecell[c]{Put\\Bottle} & \makecell[c]{Insert\\Hole} & \makecell[c]{Insert\\HDMI} & \makecell[c]{Insert\\Tube} & \makecell[c]{Grasp\\Classify} & \textbf{Avg.} \\
\midrule
ACT & 42 & 28 & 20 & 28 & 19 & 15 & 45 & 50 & 30.9 \\
VITaL & 72 & 47 & 8 & 32 & 25 & 6 & 34 & \textbf{100} & 40.5 \\
$\pi_{0.5}$ & 92 & 36 & 59 & 31 & 26 & 3 & 34 & 48 & 41.1 \\
FTP-pi0.5 & 77 & 30 & 26 & 19 & 47 & 4 & 72 & 97 & 46.5 \\
UniVTAC-ACT & 71 & 46 & 29 & 31 & 24 & 28 & 56 & 99 & 48.0 \\
FTP-1 & 97 & 48 & 65 & 47 & 64 & 3 & 79 & \textbf{100} & 62.9 \\
SimpleTouch & \textbf{100} & 62 & 68 & 50 & 85 & 3 & \textbf{100} & 99 & 70.9 \\
\midrule
Astra (V) & \textbf{100} & 30 & 80 & \textbf{90} & 60 & 50 & \textbf{100} & \textbf{100} & 76.3 \\
Astra (V+T) & \textbf{100} & \textbf{90} & \textbf{100} & \textbf{90} & \textbf{100} & \textbf{80} & 90 & \textbf{100} & \textbf{93.8} \\
\bottomrule
\end{tabularx}

%% file: table-openvitac.tex
\begin{tabularx}{\linewidth}{@{}l*{12}{C}@{}}
\toprule
\textbf{Model} & {\fontsize{6.7}{9}\selectfont\makecell[c]{Weight\\Class.}} & {\fontsize{6.7}{9}\selectfont\makecell[c]{Hardness\\Class.}} & {\fontsize{6.7}{9}\selectfont\makecell[c]{Roughness\\Class.}} & {\fontsize{6.7}{9}\selectfont\makecell[c]{Roughness\\Regrasp}} & {\fontsize{6.7}{9}\selectfont\makecell[c]{Empty-Can\\Selection}} & {\fontsize{6.7}{9}\selectfont\makecell[c]{Grasp\\Chip}} & {\fontsize{6.7}{9}\selectfont\makecell[c]{Gear\\Assembly}} & {\fontsize{6.7}{9}\selectfont\makecell[c]{Pull\\Drawer}} & {\fontsize{6.7}{9}\selectfont\makecell[c]{Insert\\USB}} & {\fontsize{6.7}{9}\selectfont\makecell[c]{Insert\\Block v1}} & {\fontsize{6.7}{9}\selectfont\makecell[c]{Insert\\Block v2}} & \textbf{Avg.} \\
\midrule
StarVLA (Qwen3) & 44 & 38 & 42 & 35 & 32 & 10 & 26 & 35 & 4 & 0 & 0 & 24.2 \\
StarVLA (Cosmos) & 51 & 67 & 94 & 9 & 13 & 1 & 33 & 22 & 3 & 1 & 3.3 & 27.0 \\
Lingbot-VA & 50 & 44 & 39 & 40 & 44 & 22 & 59 & 15 & 2 & 2 & 6 & 29.4 \\
FastWAM & 49 & 84 & 67 & 7 & 9 & 50 & 51 & 88 & 6 & 0 & 0 & 37.4 \\
XVLA & 28 & 30 & 56 & 47 & 45 & 42 & 84 & 76 & 2 & 3 & 0.7 & 37.6 \\
Xiaomi-Robotics & 52 & 46 & 54 & 62 & 38 & 24 & 73 & 84 & 5 & 2 & 0 & 40.0 \\
Lingbot-VLA-2 & 52 & 58 & 58 & 53 & 52 & 41 & 78 & 92 & 8 & 5 & 2.7 & 45.4 \\
$\pi_{0.5}$ & 53 & 60 & 64 & 52 & 48 & 39 & 87 & 99 & 14 & 9 & 9.3 & 48.6 \\
InternVLA-A1.5 & 53 & 76 & 65 & 77 & 47 & 53 & 79 & 98 & 3 & 6 & 9.3 & 51.5 \\
FTP-1 & 61 & \textbf{100} & \textbf{100} & 78 & 50 & 60 & 83 & 94 & 27 & 12 & 8.3 & 61.2 \\
OpenVTLA & \textbf{89} & 99 & \textbf{100} & 90 & \textbf{73} & 69 & \textbf{91} & 99 & 27 & 11 & 8 & 68.7 \\
\midrule
Astra (V) & 50 & 70 & \textbf{100} & \textbf{100} & 70 & \textbf{100} & 80 & \textbf{100} & \textbf{100} & 40 & \textbf{76.7} & 80.6 \\
Astra (V+T) & 40 & 90 & \textbf{100} & \textbf{100} & 50 & \textbf{100} & 80 & \textbf{100} & \textbf{100} & \textbf{70} & 70 & \textbf{81.8} \\
\bottomrule
\end{tabularx}

%% file: table-configuration.tex
\begin{tabularx}{\linewidth}{@{}l*{4}{C}l@{}}
\toprule
\textbf{Task} & \textbf{Shots} & \textbf{Segments} & $\boldsymbol{H}$ & \textbf{Total actions} & \textbf{Cameras} \\
\midrule
\multicolumn{6}{@{}l}{\textbf{UniVTAC}} \\
Lift Bottle & 1 & 12 & 10 & 120 & Fixed \\
Pull-out Key & 1 & 12 & 5 & 60 & Fixed \\
Lift Can & 1 & 12 & 5 & 60 & Fixed + wrist \\
Put Bottle & 1 & 12 & 10 & 120 & Fixed \\
Insert Hole & 1 & 12 & 5 & 60 & Fixed \\
Insert HDMI & 1 & 12 & 5 & 60 & Fixed \\
Insert Tube & 1 & 12 & 5 & 60 & Fixed + wrist \\
Grasp Classify & 2 & 5 & 4 & 40 & Fixed \\
\midrule
\multicolumn{6}{@{}l}{\textbf{OpenViTac}} \\
Weight Classify & 2 & 5 & 5 & 50 & Fixed + wrist \\
Hardness Classify & 2 & 5 & 5 & 50 & Fixed + wrist \\
Roughness Classify & 2 & 5 & 5 & 50 & Fixed + wrist \\
Roughness-Guided Regrasp & 2 & 5 & 5 & 50 & Fixed + wrist \\
Empty-Can Selection & 2 & 5 & 10 & 100 & Fixed + wrist \\
Grasp Chip & 1 & 12 & 10 & 120 & Fixed + wrist \\
Gear Assembly & 1 & 12 & 10 & 120 & Fixed + wrist \\
Pull Drawer & 1 & 12 & 5 & 60 & Fixed + wrist \\
Insert USB & 1 & 12 & 5 & 60 & Fixed + wrist \\
Insert Block v1 & 1 & 12 & 10 & 120 & Fixed + wrist \\
Insert Block v2 & 3 & 5 & 10 & 150 & Fixed + wrist \\
\bottomrule
\end{tabularx}